\pdfoutput=1

\documentclass[11pt]{article}

\usepackage{acl}

\usepackage{times}
\usepackage{latexsym}
\usepackage{booktabs}
\usepackage{graphicx}
\usepackage{multirow}
\usepackage{hyperref}
\usepackage[nameinlink]{cleveref}

\usepackage[T1]{fontenc}

\usepackage[utf8]{inputenc}

\usepackage{microtype}

\usepackage{todonotes}

\title{\dataset{}: Generating Questions to Summarize Factual Changes}

\author{Jeremy R. Cole$^1$\thanks{\enskip Equal Contribution.} \qquad Palak Jain$^1$\footnotemark[1] \qquad Julian Martin Eisenschlos$^1$ \\ \qquad \textbf{Michael J.Q. Zhang}$^3$ \qquad \textbf{Eunsol Choi}$^{3}$ \qquad \textbf{Bhuwan Dhingra}$^{1,2}$ \\
$^1$ Google Research  \qquad  $^2$ Duke University  \qquad $^3$ The University of Texas at Austin \\
\texttt{\{jrcole,palakj,eisenjulian,bdhingra\}@google.com} \\ \texttt{\{mjqzhang,eunsol\}@utexas.edu}}

\newcommand{\dataset}{\textsc{DiffQG}}
\newcommand{\qg}{\textsc{qg}}

\newcommand{\vitaminc}{\textsc{VitaminC}}
\newcommand{\wikiatomicedits}{\textsc{WikiAtomicEdits}}
\newcommand{\nlp}{\textsc{nlp}}

\begin{document}
\maketitle
\begin{abstract}
Identifying the difference between two versions of the same article is useful to update knowledge bases and to understand how articles evolve. 
Paired texts occur naturally in diverse situations: reporters write similar news stories and maintainers of authoritative websites must keep their information up to date. 
We propose representing factual changes between paired documents as question-answer pairs, where the answer to the same question differs between two versions. We find that question-answer pairs can flexibly and concisely capture the updated contents. Provided with paired documents, annotators identify questions that are answered by one passage but answered differently or cannot be answered by the other. We release \dataset{} which consists of 759 QA pairs and 1153 examples of paired passages with no factual change. These questions are intended to be both unambiguous and information-seeking and involve complex edits, pushing beyond the capabilities of current question generation and factual change detection systems. Our dataset summarizes the changes between two versions of the document as questions and answers, studying automatic update summarization in a novel way. 

\end{abstract}

\section{Introduction}

Given a pair of statements, how can we identify the difference in their information content? 
This problem has existed in different forms across \nlp{} research, such as \emph{recognizing textual entailment}~ \cite{dagan-etal-2010-recognizing} and
\emph{natural language inference}~\cite{bowman-etal-2015-large}. The initial focus of this type of research was finding the logical implication relations between sentences.

\begin{figure}[t]
\includegraphics[width=\linewidth]{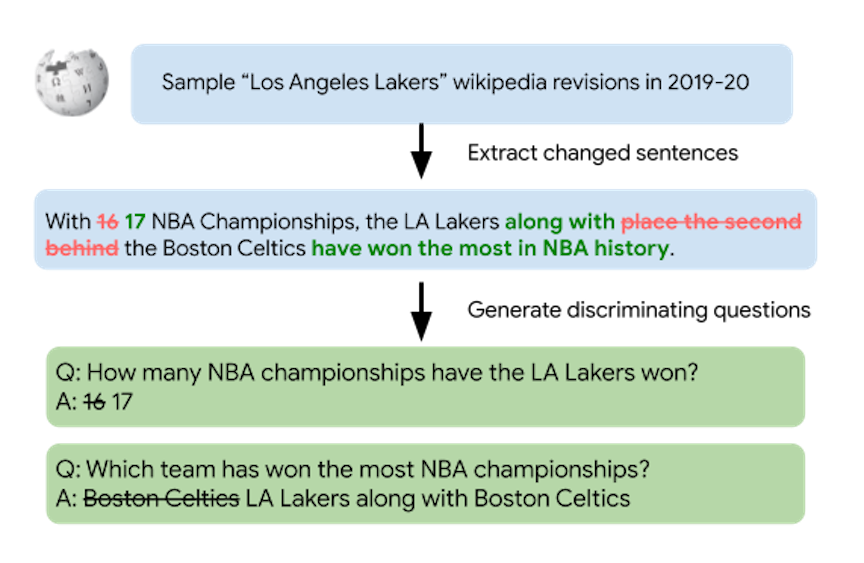}
\caption{\dataset{} consists of paired Wikipedia passages that correspond to factual edits. The goal is to generate a discriminating question given an answer span such that the question is answerable by  one of the passages but not the other or yields different answers.}
\label{fig:diffqg}
\end{figure}

More recently, specialized entailment-like resources and models have been applied to fact verification~\citep{ws-2018-fact} with applications to science, education and journalism. This trend has exposed the limited transfer
between logical entailment and general factual change detection~\citep{thorne-etal-2018-fever} as well as the need for interpretable models for this task~\citep{kumar-talukdar-2020-nile}.

Wikipedia revisions across time provide a large scale and highly available source of sentence pairs, leading to new resources such as \wikiatomicedits{} \cite{faruqui-etal-2018-wikiatomicedits} and \vitaminc{} \cite{schuster-etal-2021-get}. However, prior work is limited to minimal changes that concern only a single factual addition or change.
We introduce \dataset{}, a manually annotated dataset spanning changes over multiple years. \dataset{} consists of paired passages with complex factual changes including multiple additions and deletions within the same example. 
Additionally, it provides a way to interpret the prediction in the form of a discriminative question-answer pair that identifies the change.

Question-answer pairs provide a semi-structured summary of a change: more flexible than knowledge graph triples and more useful than free-form text. 
%
%
For instance, question-answer pairs can represent different types of updates: a new prime minister may update an answer, while a new type of minister would add an entirely new question. 
%
%
%

Question generation (\qg{}) is a new \nlp{} task that consists of generating a question that a provided document answers. There are various successful applications of this approach, including augmenting datasets to train question answering systems \citep{duan-etal-2017-question, lewis-etal-2021-paq}, capturing implicit information written about text \citep{pyatkin-etal-2021-asking}, and building soft knowledge bases \citep{chen-etal-2022-qamat}. Previous work in \qg{} treated the underlying passages as static \citep{lewis-etal-2021-paq}, while real life documents are constantly updated~\citep{dhingra-etal-2022-time}. As the source corpus is updated, new question-answer pairs must be added and existing ones must be updated. 

\dataset{} thus 
addresses two challenges simultaneously: providing an interpretable summarization of factual changes and updating soft knowledge bases consisting of question answer pairs. 
We hope that this dataset can also help evaluate the quality of \qg{} models in producing natural, semantically \ correct, unambiguous, and information-seeking questions.
The dataset and code for our experiments will be open sourced.\footnote{\url{https://github.com/google-research/language/tree/master/language/diffqg}}

Our contributions are the following:

\noindent \textbf{(a)} We introduce \dataset{}, an expert-annotated \textit{evaluation} dataset that consists of questions that summarize the difference between two passages. To the best of our knowledge, no prior dataset exists that covers such long and complicated edits.


\noindent \textbf{(b)} We propose a set of metrics that can be used to measure improvements in question generation or factual change detection. 

\noindent \textbf{(c)} We evaluate a comprehensive set of baselines that surface the shortcomings of current systems.

\section{\dataset{} Task}


The goal of \dataset{} is to capture how two similar passages differ from each other using question-answer pairs. 
In particular, given a base passage $x_b$ and a target passage $x_t$, where $x_t$ and $x_b$ are different versions of the same article, we aim  to generate discriminating questions $Q_t$. For each $q_t \in Q_t$, the information to deduce the corresponding answer span $a_t \in A_t$ must be missing in $x_b$. To limit the scope of possible questions, each answer span $a_t \in A_t$ must be a substring of $x_t$. While $a_t$ could also be a substring of $x_b$, $x_b$ must be missing the required information to deduce $a_t$ is the correct answer. Alternatively, there could be a corresponding answer span $a_b$, which is the answer resulting from answering $q_t$ with $x_b$. Note that we consider paraphases of $a_t$, such as lexicalizing numbers and using alternate entity names, as equivalent answers. 

This discriminating question has certain additional requirements: it should be  seeking factual information and stand-alone~\cite{choi-etal-2021-decontextualization} (i.e., interpretable when presented by itself without the passage). It is possible that no such discriminating question can be written. The annotators only mark that there is no factual change when they are fairly confident that there is no new information about the answer span in the target passage.

Consider the following example:

\begin{itemize}
 \item $x_b $ = John Doe won two gold medals at the Olympics in 2012.
 \item $x_t $ = John Doe won a gold medal at the Olympics in 2012.
\end{itemize}



Annotators are informed that the goal of the process is to collect \textit{disambiguated} and \textit{information-seeking} queries that can be answered with one passage but not with the other. By \textit{disambiguated} queries, we mean queries that refer to roughly a single answer without any context. For instance, ``Who won two gold medals in the 2012 Olympics?'' could refer to several different people, and questions of the form ``How many medals did he win in the 2012 Olympics?'' are not answerable at all without the presence of the John Doe passage.

Information-seeking queries are ones where the questioner would not need to know the answer in advance for the question to make sense. This is related to the original goals of Natural Questions \citep{kwiatkowski-etal-2019-natural} and corresponds to the \emph{Cranfield}-style questions described by \citet{rodriguez-boyd-graber-2021-evaluation}. As an example, ``What did Al Capone's mother do for a living?'' seems like an information-seeking query. On the other hand, ``Which Italian-American gangster's mother was a seamstress?'' does not: why would the questioner assume that such a person even exists unless they already knew the answer?
We describe the annotation process to acquire such discriminating question set in the next section.

\section{Data Collection}
\label{Data Collection}
\begin{figure*}[t]
\includegraphics[width=\textwidth]{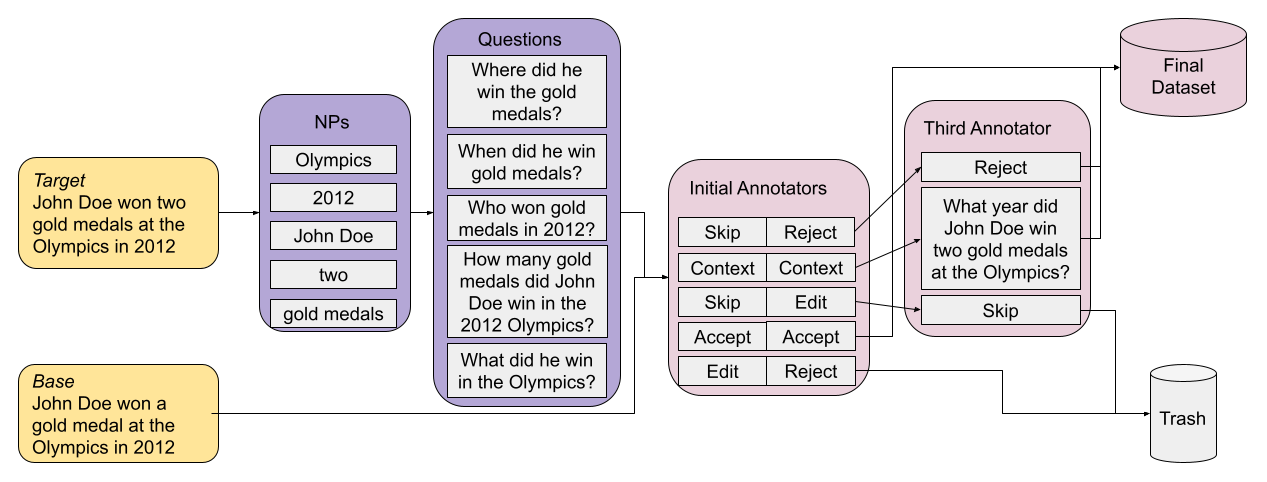}
\caption{\dataset{} annotation process. Noun phrases are extracted from the target passage and a question generation model seeds initial questions. Annotators decide if the generated questions serve as satisfactory discriminating questions (Accept), must be edited (Context, Edit), or contain no new factual information (Reject). After the first phase, a third annotator resolves indecision (Skip), leaving us with a set of questions and negative examples. If the original two annotators disagree or the third annotator cannot resolve indecision, the example is discarded.}
\label{fig:data}
\end{figure*}

Collecting such discriminating questions is a non-trivial process. Thus, we introduce a staged annotation process with expert annotators (the authors of this paper) and use a question generation model to aid annotation. We describe our process below (visualized in \autoref{fig:data}).



\subsection{Input Passage Pair Selection}\label{sec:initial_data}
First, we extract the Wikipedia pages for entities from the Natural Questions (NQ) training set~\citep{kwiatkowski-etal-2019-natural}. In particular, we find the pages for Wikipedia snapshots between the years 2008 and 2020. After sampling a base document, we find the version of that document one year later and use this as the target document. Using the two documents as a corpus, we compute cosine similarity between the TF-IDF vectors over each sentence pair and pick the pair with the highest similarity.
Sentence pairs with similarity either greater than $0.8$ or less than $0.25$ are discarded. In order to retain meaningful changes, we ensure at least one noun or number is edited and up-sample instances where either a named entity or at least five tokens have been edited. 


This process thus focuses on edits accumulated over a year and consists of changes ranging from five to twenty tokens, making these semantically richer and more widely applicable than existing factual change detection datasets.

\subsection{Seed QA Pair Generation}\label{sec:seed}
Each target passage has a very large number of possible answer spans; for convenience, we restrict them to only noun phrases identified using the Berkeley Neural Parser \citep{kitaev-klein-2018-constituency}. 
To increase annotation speed, each example starts with a \textit{seed question} that is generated by a question generation model from the target passage and answer span. In particular, we use a T5-XXL model \citep{raffel2020exploring} that has been finetuned on the SQuAD dataset \cite{rajpurkar-etal-2016-Squad}.

\subsection{Annotation Process}
\dataset{} annotation was done in three phases by six expert volunteers. First, annotators are given the paired passages described above along with the answer span and seed question, which corresponds to one \textit{example candidate}. Then, they label each example candidate with one of the five options:

\paragraph{Accept} The seed question follows all requirements for discriminating questions as is.
\paragraph{Context} The seed question asks about the appropriate topic but is not answerable outside of the context of the passage. For instance, questions like ``What did he win?'' or ``Where were the Olympics held?'' both lack context in order to answer the question successfully. 
\paragraph{Edit} The example candidate answer has a discriminating question, but the question is different than the seed question. Sometimes, this is because the seed question does not capture the new information contained in the passage; other times, the seed question is simply nonsense. 
\paragraph{Reject} This example candidate has no valid discriminating question. In other words, these are negative examples. Sometimes, the target passage contains no new information at all; however, it may contain new information about other answer spans but not the one in the example candidate. In our previous John Doe example, there is no new information about ``the Olympics'', except indirectly.
\paragraph{Skip} It is unclear if there is a valid discriminating question for this example candidate. This could be due to awkward or cumbersome answer spans: for instance ``two gold medals at the Olympics in 2012.'' Alternatively, it could seem unclear if there is new information about an answer span due to its indirect relationships with other entities. Finally, it could be difficult to write an information-seeking question even though there is obviously new information: for instance, writing a question with the answer span ``John Doe'' in the previous example.

Each example candidate is considered by two annotators. Unless both annotators agree to Add, Reject, or Skip, a third annotator decides. In examples where one annotator chose Context or Edit, the third annotator is responsible for writing the correct question according to the guidelines. If one annotator chose Add or Reject and the other skipped, the third annotator can confirm the Add or Reject or also skip if they cannot decide. See \Cref{sec:interface} for the annotation interface.

\subsection{Question Writing Guidelines}
Note that writing a single, context-free, and information-seeking question that summarizes the difference between the two passages can be challenging. In cases where it seemed impossible, annotators are encouraged to skip the example. For cases where additional Context was needed, annotators are encouraged to add as much context without sacrificing fluency, so that the question can be answered without awareness of the source passage. When an annotator writes a question from scratch in the Edit case, they are encouraged to think of a question that either would have a different answer, be unanswerable, or have a false precondition if posed against the base passage. While conditioning on a single answer span reduces ambiguity, the task is still ambiguous, which is unavoidable when handling large and complex edits. 
 
 \begin{table}[htb!]
\begin{center}
\small
\begin{tabular}{lccc}
\hline

& Passages & Answers & Avg. Edited tokens \\
\hline
w/ change & 391 & 759 & 12.9 \\ 
w/o change & 478 & 1153 & 14.1 \\
Total & 672 & 1912 & 13.7  \\ 
\hline
\end{tabular}

\caption{Dataset statistics for \dataset{}. Edited tokens represents the average tokens added or removed in a given passage pair.}
\label{tab:data}

\end{center}
\end{table}

\subsection{Data Statistics}
Our initial annotation process starts with $8,530$ example candidates drawn from $999$ passage pairs. Annotators skipped nearly $75\%$ of the example candidates, leaving $1,912$ examples. Of those, roughly $40\%$, or $759$, had a factual change and thus a discriminating question written about them, leaving $1153$ negative examples.
%
Of the spans where a factual change was detected, annotators  modified the question in $65\%$, or $494$, of the examples: $45\%$ are labeled as Context and $20\%$ as Edit. Detailed dataset statistics can be found in \autoref{tab:data}.

Note that on all cases where a question was accepted as is or considered a negative example, at least two annotators agreed on that rating. However, human written questions are not verified; both annotators agree that there exists a discriminating question but not necessarily what it is. To address this, we evaluate a small set of fifty questions and found that a second annotator would write an equivalent question around 85\% of the time.

\section{Motivation}
In the previous section, we described \dataset{} and its annotation procedure. As mentioned, the purpose of \dataset{} is to detect and describe factual changes. In particular, \dataset{} is a rough measurement of a model's ability to automatically construct a database of question-answer pairs that encapsulate the changes. There are many possible formats that could be used as an alternative to summarize factual changes, such as paragraphs, knowledge base triples, or individual claims.

While paragraphs can contain nuance, they lack atomicity. It is thus difficult to tell what exactly changed or otherwise compare two changes to each other. This makes them less useful as a database. 

On the other hand, knowledge base triples are limiting in the types of factual changes that can be described: regardless of the exact setup, the nodes and relations come from some form of fixed vocabulary that may require discarding interesting changes. For instance, changes related to a set of entities, date ranges, various numbers, or abstract information may all be challenging.

Another alternative method would be a list of claims, similar to Vitamin-C \citep{schuster-etal-2021-get}. This method is also atomic and more flexible than knowledge base triples. However, question-answer pairs have a few advantages. First, question-answer pairs are semi-structured information, forming a loose key-value pair. Factual edits may change the answer to an existing question or add information corresponding to an entirely new information, requiring a new question. Conversely, claims are more difficult to relate to each other.

Finally, question-answer pairs are interesting because question answering is interesting. Previous work has seen the use of a database of question-answer pairs as a method to improve question answering performance \citep{lewis-etal-2021-paq}. A good method for automatically creating and updating such a database thus seems quite useful. As factual corpora change over time, we envision constructing such a database to require iterative updates. 

\section{Metrics}\label{sec:metrics}
\dataset{} can be used to measure performance on three related tasks. 

\paragraph{Factual Change Detection} \label{Metric:Diff Detection}  Given an example consisting of a base passage, target passage, and answer span, the goal is to determine whether there exists a valid differentiating question. In other words, whether there is new information about this answer span that is present in the target passage when compared to the base passage. To measure this, we report accuracy, precision, recall and F1 score over the existence of a differentiating question in our annotations. Note that always predicting no change achieves 60.3\% accuracy but 0\% F1, but random guessing corresponds to 44.1\% F1. 

\paragraph{Discriminating Question Generation} Given a target passage and answer span, write a specific, unambiguous and information-seeking query that can be answered  with the target passage. To measure this, we compare machine generated questions to those that humans verified, edited, or hand wrote. We use two model-free metrics Rouge-1 and Rouge-L \citep{lin2004rouge} which measure the token-level overlap and longest subsequence overlap of the questions, respectively. We also consider two model-based metrics, BLEURT \citep{sellam-etal-2020-learning}, which is a learned evaluation for text similarity based on BERT \citep{devlin-etal-2019-bert}, and a query similarity model \citep{reimers-2019-sentence-bert} trained on Quora Question Pairs \footnote{\href{https://huggingface.co/cross-encoder/quora-roberta-large}{huggingface.co/cross-encoder/quora-roberta-large}}.

Note that we evaluate discriminating question generation despite using a question generation model in our annotation procedure. Note that all of these questions are reviewed by humans and only the very fluent ones are kept. As question generation models vary in which of their productions are very fluent, this set is less trivial than it would initially appear. Nonetheless, we also separate human-written or edited questions and evaluate that set independently. 

\paragraph{Full System} This is the overall measure of performance on \dataset{}. We reuse the metrics from discriminating question generation, using 0.0 for BLEURT, ROUGE-1, ROUGE-L, and Query Similarity if the factual change detection is incorrect.  
\section{Methods}\label{sec:methods}
\begin{figure*}[t]
\includegraphics[width=\textwidth]{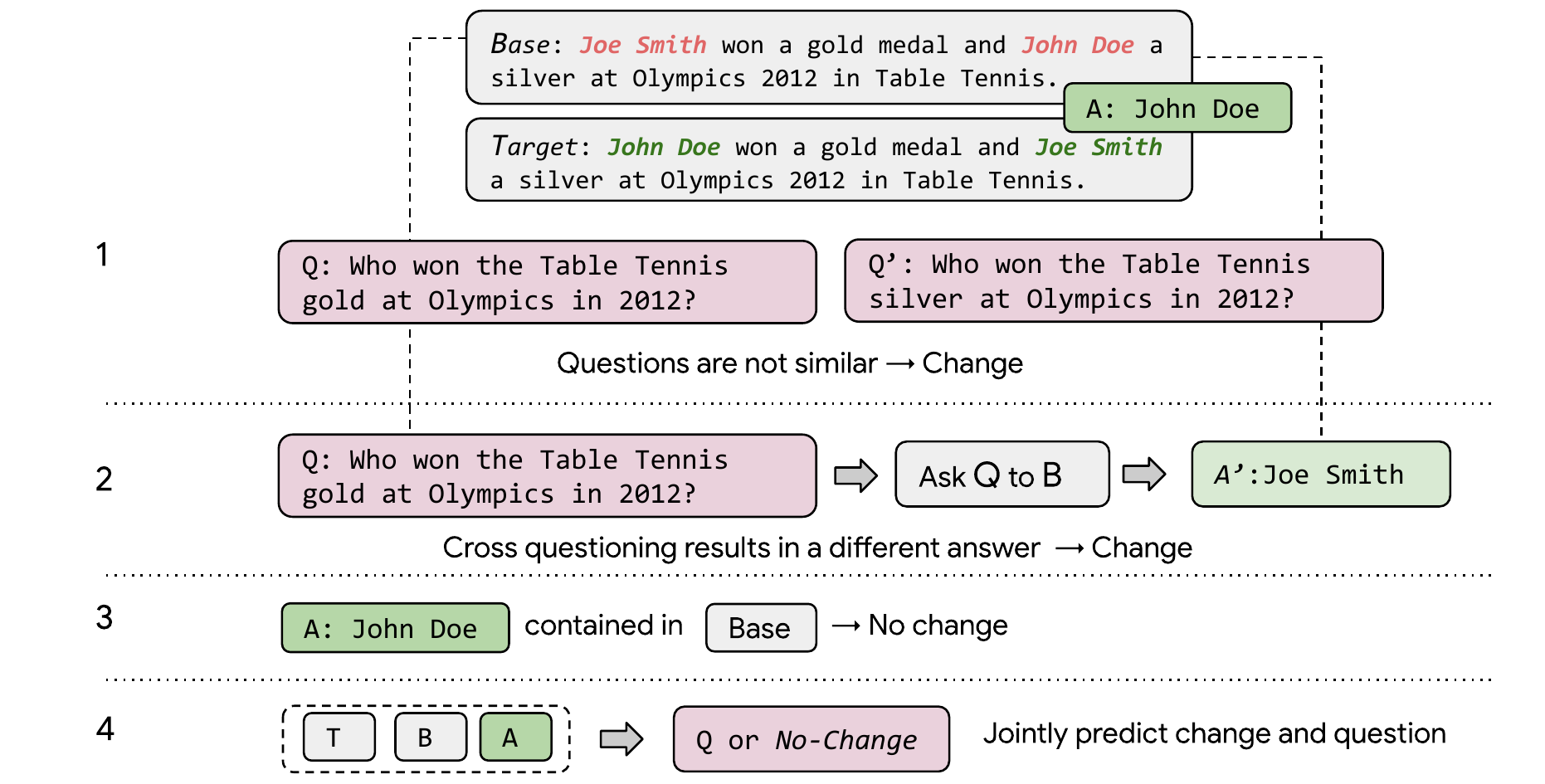}
\caption{Methods for factual change detection on \dataset{}: (1) Question similarity (2) Cross-Questioning.  The
QA-equivalence method combines (1) and (2), deciding it is a Change only when both the systems find a Change. (3) Overlapping Answers (4) Language model  that jointly learns the task of factual change detection and discriminating question generation, decoding the discriminating question or a special token indicating no change.}
\label{fig:method}
\end{figure*}

As mentioned, \dataset{} can be thought of as a composition of two tasks: factual change detection and discriminating question generation.
Our simple baseline systems thus treat this as a pipeline, first predicting whether or not there is a factual change and then generating a discriminating question if there is. We also present baseline models that solve both tasks jointly with a single prediction. 
Our methods are illustrated in \autoref{fig:method}.

Note that none of our methods use any part of \dataset{} as training data, as the dataset is only intended to be used for evaluation. Models are instead trained on larger existing datasets for question generation and factual change detection. 


\subsection{Factual Change Detection}\label{subsec:fact}
We propose five baselines based on answer equivalence or both question and answer equivalence. 

\subsubsection*{Answer Equivalence Baselines}
\label{Method:Overlap}
Our trivial baseline (\textit{Overlapping Answer}) classifies an example as having a factual change if and only if the answer span is not present in the base passage. The span is normalized before looking for token overlap with the passage. 

Our simple model-based baseline is similar but uses an Answer Equivalence model \citep{bulian2022tomayto}. It compares the target answer span against all valid base answer spans, finding a factual change if it does not match any of them. The Answer Equivalence model additionally takes as input a candidate question for each answer span.

\subsubsection*{Question-Answer Equivalence Baselines}
\label{Method:Query Similarity}
The previous methods only consider the answer span, ignoring the context. Here, we consider methods that also use a question generation model on the passages and answer span to determine if there is new information.

For the first method, we find base answer spans equivalent to the target answer span using the \textit{Overlapping Answer} method. Then, we want to see if the questions generated from those answers would also be equivalent in both passages. To do so, 
we use a T5-XXL \citep{raffel2020exploring} model trained on Quora Question Pairs \footnote{https://www.kaggle.com/c/quora-question-pairs} to predict whether the pair of questions is ``duplicate'' or ``not duplicate'' . If the question is not a duplicate, then we consider this example to have a factual change. Thus, answer spans present in both passages but with different contexts could now be identified as having a factual change. This will increase the recall of the \textit{Overlapping Answer} method.


The second approach adds a cross questioning filter (\textit{Cross-Q}). Given a candidate question generated from target passage, we attempt to answer the question with the base passage
using a reading comprehension model. We train a T5-XXL model  on SQuaD v2 \citep{rajpurkar-etal-2018-know} 
question-answering dataset to take the passage and question as input and output the answer. If the model predicts no answer or a different answer from the target span, we classify the example as having a factual change. 
Finally, the \textit{QA-equivalence} method combines both the query similarity model and cross question model to boost precision. In this case, we consider an example to have a factual change only when both methods determine a factual change.
\label{Method:Cross Question Answering}

\subsection{Question Generation}
\label{Method:Question Generation}
Each of our factual change detection baselines is then combined with a question generation model. We use a similar T5-XXL model finetuned on SQuAD as described in \Cref{sec:seed}.
Unsurprisingly, the model we use to seed the questions can do well on the questions that it wrote originally; however, this is an unfair baseline. Thus, we additionally test a version of the model that is sampled and also a retrained version using a different seed. We also test training a similar model trained on Natural Questions \citep{kwiatkowski-etal-2019-natural}.

\subsection{Joint systems}
Many of the techniques described \Cref{subsec:fact} are inefficient, requiring multiple runs of various models. For instance, the Query Similarity method requires one model run for each answer span in the base passage per example, which corresponds to quadratic runs for each pair of passages. We also explore methods that can directly compare the base and target passage without the need for any intermediate steps. These methods instead jointly detect if there is a factual change and generate a discriminating question.

\subsubsection*{Finetuning on Silver Data}
\label{Method:Finetuning on QA-equivalencea}
We mine additional pairs of Wikipedia passages using the same process as in \Cref{sec:initial_data}. We then identify every possible answer span from the target passage. We create silver training examples for the factual change detection component of the task by labeling each target answer span using our best heuristic method, \textit{QA-equivalence}.

We then convert these labels into a text-to-text task. For each example with a factual change, we use the question generated by the SQuAD model (\Cref{Method:Question Generation}) as the target. For questions without a factual change, we use ``None'' as the target. The input to the model is the concatenation of the base and target passages with the target answer span marked by a special token.

The model is a T5-XXL initialized from the same question generation model as the model that produced the original questions.


\subsubsection*{Finetuning on \vitaminc{}}
The \vitaminc{} task \cite{schuster-etal-2021-get} also has a factual change detection component. We sample negatives from the \vitaminc{} Revision Flagging dataset, using negative examples with a random noun phrase chosen as the answer span.

For positive examples, we need to identify a specific answer span that contains a factual change as well as the corresponding discriminating question. While there is no direct counterpart of this task in \vitaminc{}, the Fact Verification task is somewhat similar. The dataset consists of an evidence \emph{e}, a simplified claim \emph{c} supporting \emph{e}, a companion edited sentence \emph{e'} and an edited-claim \emph{c'} refuting \emph{e} and supporting \emph{e'}. An answer span \emph{a} is identified based on the token-level diff between \emph{(c, c')} and a question generated from \emph{c} conditioned on \emph{a} using the question generation model in \Cref{Method:Question Generation}. Because \emph{c} is a simple sentence, we anecdotally find the generated questions to be of high quality.

The dataset (\emph{e}, \emph{e'}, \emph{a}) is converted to a text-to-text task and used to finetune a T5-XXL model following the same steps as above.
Note that a model trained on an equal amount of positives and negatives yielded poor performance on \dataset{}. In our final \vitaminc{} silver dataset, we used only 10\%  negatives to achieve a reasonable performance.
\section{Results and Discussion}
We present results separately for factual change detection, question generation and the full system. We also report results separately for the overall performance and the performance on only the subset of questions that are human written; those sentences labeled as Edit or Context in the annotation phase. Selected examples with model outputs are provided in \Cref{sec:appendix:failure} to illustrate the capabilities and typical errors baseline.

\subsection{Factual Change Detection}
\label{subsec:results:fact}
\begin{table}[t]
\begin{center}
\scalebox{0.9}{
\begin{tabular}{lcccc}
\toprule
Model & Acc & P & R & F1 \\
\midrule
Random & 50.0 & 39.5 & 50.0 & 44.1 \\
Overlapping Ans & 82.1 & 79.1 & 74.6 & 76.7 \\ 
Answer Equivalence & 81.1 & 84.6 &	64.2 &	73.0 \\
Query Similarity & 77.7 & 65.4 & 93.3 & 76.9 \\
Cross-Q & 76.9 &	65.9 &	86.8 & 74.9 \\
QA-equivalence & \textbf{83.9} & 76.7 & 85.5 & \textbf{80.9} \\
FT on QA-equivalence & \underline{82.5} & 78.4 & 77.1 & \underline{77.7} \\ 
FT on \vitaminc{} & 81.5 & 79.9 & 71.4 & 75.4 \\
\bottomrule
\end{tabular}
}
\caption{Metrics for factual change detection. Note that none of these models have change detection training data and are instead verifying with other tasks or heuristics. The random baseline assumes guessing change or No change with equal probability. 
Acc=Accuracy,P=Precision,R=Recall,F1=F1 Score. \textbf{Bold} indicates the best model, second best model is \underline{underlined}.}
\label{tab:diff}
\end{center}
\end{table}

\autoref{tab:diff} compares the performance of various systems on the factual change detection task.
We find that QA-equivalence performs better than heuristic baseline methods. In particular, it better handles cases where the answer span text is unchanged, but the surrounding context has changed. For example, in the passage \textit{``On the New Hampshire Executive Council, Laconia is in the 1st District, represented by <ADD: Republican Joe Kenney> <DEL: Democrat Michael J. Cryans>."}, QA-equivalence correctly captures the new information associated with the answer span \textit{``New Hampshire Executive Council"} in the form of the question \textit{``What state council does Joe Kenney represent Laconia in?"} 					
However, the method is prone to detecting spurious changes even when the passages have no semantic edit as illustrated in \Cref{sec:appendix:failure}.

The joint system finetuned on silver data from QA-equivalence does not seem to improve upon QA-equivalence. 
While it seemingly benefits from the additional context, it still struggles with long and complex edits. However, this model only requires a single inference to do both tasks.

The \vitaminc{} trained model, despite having access to additional data, was also unable to improve on our baseline. \vitaminc{} style edits are substantially different than \dataset{} edits, generally only consisting of small changes. Thus, the model finetuned on \vitaminc{} performs poorly on large phrase changes or sentence refactors. 

\subsection{Question Generation}
\begin{table}[htb]
\begin{center}
\scalebox{0.89}{
\begin{tabular}{lcccc}
\toprule
Model & R-1 & R-L & QSim & BLRT\\
\midrule
SQuAD-seed & 71.7 & 74.9 & 66.3 & 74.1 \\
SQuAD-sampled & 59.6 & 63.2 & 57.8 & 65.6 \\
SQuAD-retrained & 58.0 & 61.7 & 58.9 & 65.6 \\
NQ & 20.4 & 39.6 & 22.9 & 41.5 \\
\bottomrule
\end{tabular}
}
\caption{Variation in performance of question generation models on the positive subset of \dataset{}. We report three different versions of the SQuAD model, where the first model is the same as we used to seed the annotations. R-1=ROUGE-1, R-L=ROUGE-L, QSim=Query Similarity model-based accuracy, BLRT=BLEURT.}
\label{tab:qg}
\end{center}
\end{table}

In \autoref{tab:qg}, we compare our question generation baseline models on the subset of the positive examples. In \autoref{tab:qg_edited_tp}, we examine the same models on the subset of those that are human written: examples with a change from \Cref{tab:data}.

\begin{table}[ht!]
\begin{center}
\scalebox{0.89}{
\begin{tabular}{lcccc}
\toprule
Model & R-1 & R-L & QSim & BLRT\\
\midrule
SQuAD-seed & 56.5 & 61.4 & 48.2 & 61.3 \\
SQuAD-sampled & 50.9 & 55.2 & 47.0 & 58.3 \\
SQuAD-retrained & 50.3 & 54.5 & 50.4 & 59.4 \\
NQ & 20.4 & 39.1 & 20.0 & 40.2 \\
\bottomrule
\end{tabular}
}
\caption{Variation in performance of question generation models on human written questions of \dataset{}. The first model is the same as what we used to seed the questions. R-1=ROUGE-1, R-L=ROUGE-L, QSim=Query Similarity model-based accuracy, BLRT=BLEURT.}
\label{tab:qg_edited_tp}
\end{center}
\end{table}

The primary goal of this evaluation is to test whether the questions directly produced by the seed model described in \Cref{sec:seed} are still useful for evaluating systems on \dataset{}. We find from sampling from that same model and from retraining with the same process (as described in \Cref{Method:Question Generation}) that performance on the overall set degrades considerably. This suggests that unless someone had access to the same model, these questions that are human-verified but not human written can still be useful for evaluation.
Nevertheless, the seed model can be thought of as a rough ceiling on current question generation performance on \dataset{}.

The human written questions (see \autoref{tab:qg_edited_tp}) seem to be much more challenging for the question generation models to replicate. Performance degrades substantially: naturally it degrades the most for the seed model that wrote some of the questions in the overall dataset, which it should exactly match. 

We note also that a question generation model finetuned on Natural Questions  \citep{kwiatkowski-etal-2019-natural} yields a significantly different question style than SQuaD. This is likely because SQuAD questions are originally generated from passages, while Natural Questions are more free form. In addition the  Natural Questions  model is found to hallucinate in numerous scenarios.
This reflects on the poor performance of the Natural Questions-trained question generation model on \dataset{}. 

As a caveat, the possible universe of questions written to summarize a factual change can be very large. While restricting to a single answer span reduces this space, we still find scenarios with multiple valid questions. Thus, there may be some disagreements where the model generates a completely valid question that is simply not the most pertinent one according to our annotators.





\subsection{Full System}
\begin{table*}[ht]
\begin{center}

\begin{tabular}{ll|ccc|ccc|ccc}
\multicolumn{2}{c|}{\multirow{2}{*}{Models}} & \multicolumn{3}{c|}{\multirow{2}{*}{Factual Change}} & \multicolumn{6}{c}{Full System} \\
\cmidrule(lr){6-11}
\multicolumn{2}{c|}{} & \multicolumn{3}{c|}{Detection} & \multicolumn{3}{|c}{All} & \multicolumn{3}{|c}{Human written} \\
\midrule
Change & QG & P & R & F1 & R-L & Qsim & BLRT & R-L & Qsim & BLRT \\
\midrule
\emph{Pipelined systems} & \\
Overlapping Answer & SQuAD & 79.1 & 74.6 & 76.7 & 71.3 & 70.8 & \underline{72.3} & 40.6 & 38.1 & 43.6 \\
QA-equivalence  & SQuAD & 76.7 & 85.5 & \textbf{80.9} & 71.4 & 70.6 & \textbf{72.7} & 46.8 & 43.7 & \textbf{50.8} \\
\midrule
\emph{Joint systems} & \\
\multicolumn{2}{l|}{FT on QA-equivalence} & 78.4 & 77.1 & \underline{77.7} & 71.2 & 70.0 & \underline{72.3} & 42.7 & 36.6 & \underline{45.9} \\
\multicolumn{2}{l|}{FT on \vitaminc{}} & 79.9 & 71.4 & 75.4 & 69.9 & 68.1 & 71.1 & 36.3 & 32.2 & 39.9 \\
\bottomrule
\end{tabular}

\caption{Full system performance of the pipelined and joint systems on \dataset{}. Note that the ``All'' component of the full system metric includes all of \dataset{} while the ``Human written'' portion includes only questions edited by the annotators. The pipelined systems use the retrained SQuaD model for their question generation component. \textbf{Bold} represents the best system, second best is \underline{underlined}. R-L=Rouge-L, Qsim=Query similarity model based accuracy, BLRT=Bleurt, FT=finetuned}
\label{tab:overall}

\end{center}
\end{table*}


Full \dataset{} metrics are presented in \autoref{tab:overall} and include the two finetuned systems that are trained on \vitaminc{} and QA-equivalence, respectively, as well as two pipelined systems with factual change detection models attached to a question generation model. For the pipelined experiments, we use the retrained SQuaD model described in \Cref{Method:Question Generation}. We evaluate these models on the full \dataset{} as well as human written subset.  

Overall, all of the systems are relatively close in performance. QA-equivalence works the best, with the finetuned version and simple heuristic model close behind, indicating substantial room for future innovation. 
On the human written subset, the performance drops significantly further highlighting the challenge of the human written questions.


\section{Related Work}

\paragraph{Factual Edits}
Factual change detection has been of recent interest to the community. For instance, \wikiatomicedits{} \citep{faruqui-etal-2018-wikiatomicedits} rely on Wikipedia revisions to learn to discriminate factual edits. Closest to our work is \vitaminc{} \citep{schuster-etal-2021-get} which aims to generate a discriminating claim given a pair of edited sentences. However, both of these datasets primarily rely on smaller edits, frequently consisting of a single entity or number substitution. For instance, \vitaminc{} examples have a median of four token changes and \wikiatomicedits{} examples have a median of two token changes. Moreover, these edits are easier to detect using heuristics such as  noun or entity overlap. On the other hand, \dataset{} examples have a median of thirteen token changes that can involve  multiple entity updates. Further, the surrounding contextual information for an entity could be updated even when the entity itself is present in both passages. This makes \dataset{} edits harder to summarize and substantially different than previous work; this is also observed in \Cref{subsec:results:fact} where using \vitaminc{} training data to solve \dataset{} yields poor performance.

Recent work such as Fruit \citep{iv-etal-2022-fruit} and PEER \citep{schick2022peer} also operate on more complicated edits. Fruit generates updated sentences from a base passage given the new evidence in a Wikipedia article. PEER attempts to imitate the editing process using a sequence of planning steps. However, both of these primarily focus on generating the target update, while we focus on succinctly capturing the edited information. Further, the use of question generation as a device for discrimination is novel to the best of our knowledge.

\paragraph{Question Generation}
Question generation has been successfully applied to various purposes, including augmenting question answering systems \citep{duan-etal-2017-question, lewis-etal-2021-paq}, capturing implicit information written about text \citep{pyatkin-etal-2021-asking}, and building soft knowledge bases
 \citep{chen-etal-2022-qamat}. In this work, we apply question generation to the task of discriminating edited sentences. As far as we are aware, there is no prior work on evaluating question generation systems. 

\section{Conclusion}
In this work, we introduce the \dataset{} task and dataset to evaluate the ability of \nlp{} systems to summarize changes between two related passages via question generation. We present several heuristic and model baselines as well as a set of metrics to measure performance on the dataset. The \dataset{} task requires models to identify changes in factual relationships and ignore other stylistic edits. We find that existing approaches struggle under these conditions. Models trained to perform factual change detection and question generation jointly sometimes fail to understand even simple edits. We hope this work finds value in future research on this important problem.

\section*{Limitations}
\dataset{} is relatively small, consisting of less than a thousand questions and less than two thousand total examples. This makes us unable to provide a training set, limiting claims we can make about the difficulty of the task. Moreover, summarizing complex edits can have a large space of valid solutions. While using questions conditioned on an answer reduces this space, there's still room for ambiguity.

To make annotation easier, we use a question generation model; however, our goal is also to evaluate question generation models, complicating our story. Finally, most of the baselines we evaluate are some form of T5 \citep{raffel2020exploring} model. It is possible that other model architectures could have solved this task more effectively.

\section*{Acknowledgements}

We thank Jannis Bulian, Tal Schuster and William Cohen, as well as our anonymous reviewers, for their thoughful comments and valuable feedback.

\bibliography{anthology,custom}
\bibliographystyle{acl_natbib}

\clearpage
\appendix
\section{Appendix}

\subsection{Qualitative Examples}
\label{sec:appendix:failure}
\begin{figure*}[t]
\caption{\dataset{} examples with predictions from various systems. The edited sentence is color coded with green for added tokens and red for deleted; the answer span is underlined. Additional context is omitted unless required for illustration (provided in gray). \textit{No change} indicates there was no factual change for the example.}
\includegraphics[width=\textwidth]{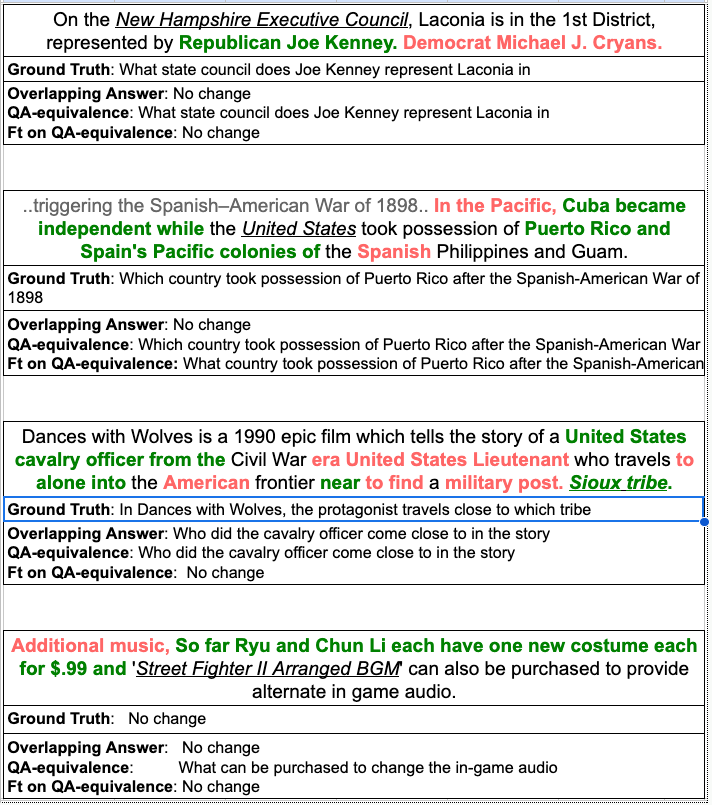}
\label{fig:examples}
\end{figure*}

Examples from DiffQG dataset are illustrated in \autoref{fig:examples}. The model outputs (success or failure) from various systems are also provided alongside. 

\subsection{Annotation Interface}
\label{sec:interface}
\begin{figure*}[t]
\includegraphics[width=\textwidth]{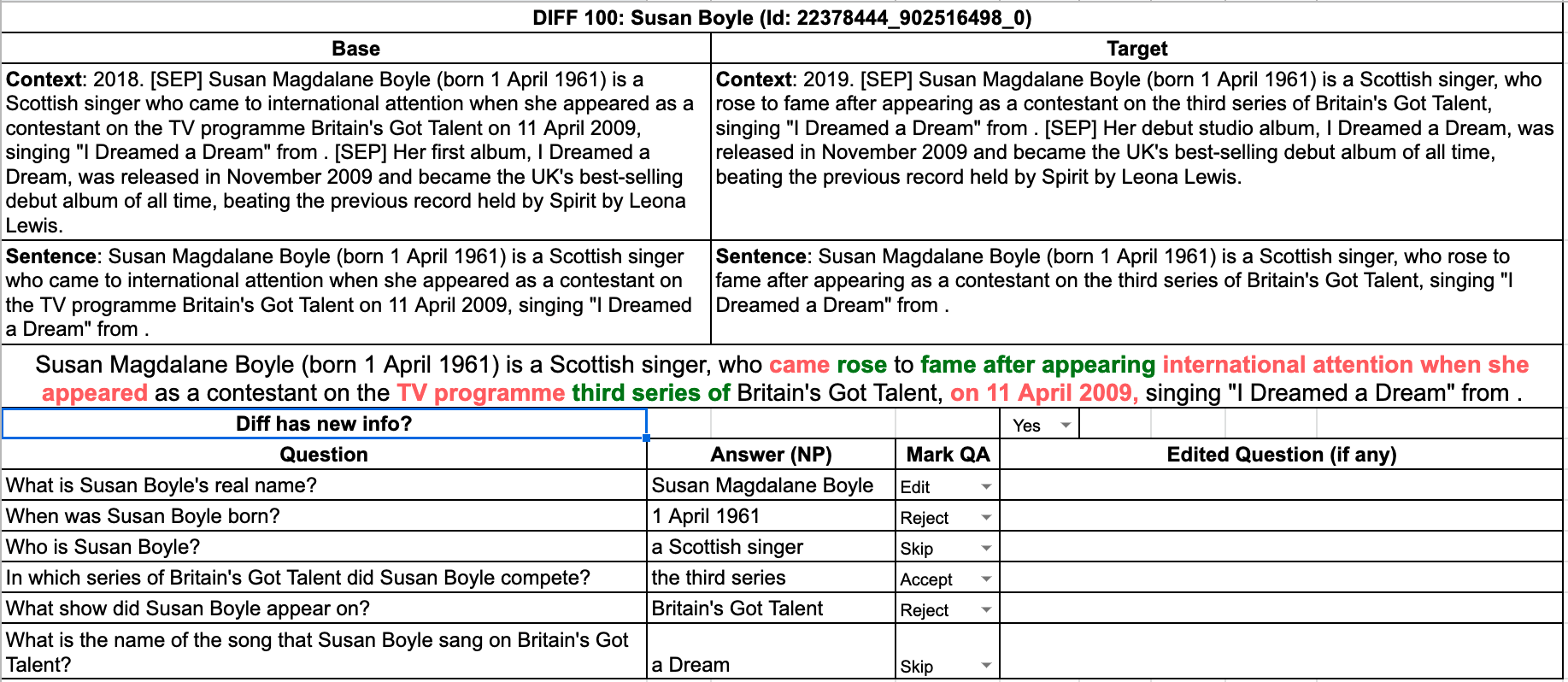}
\caption{Interface for the first phase of annotations, where an annotator chooses one of the five options: Accept/Reject/Edit/Context/Skip. Each example is annotated by two annotators. If both agree, the example is accepted as is or goes to a third annotator for editing. If one of the annotators skips, the third annotator makes the final decision.}
\label{fig:annotation_p1}
\end{figure*}

\begin{figure*}[t]
\includegraphics[width=\textwidth]{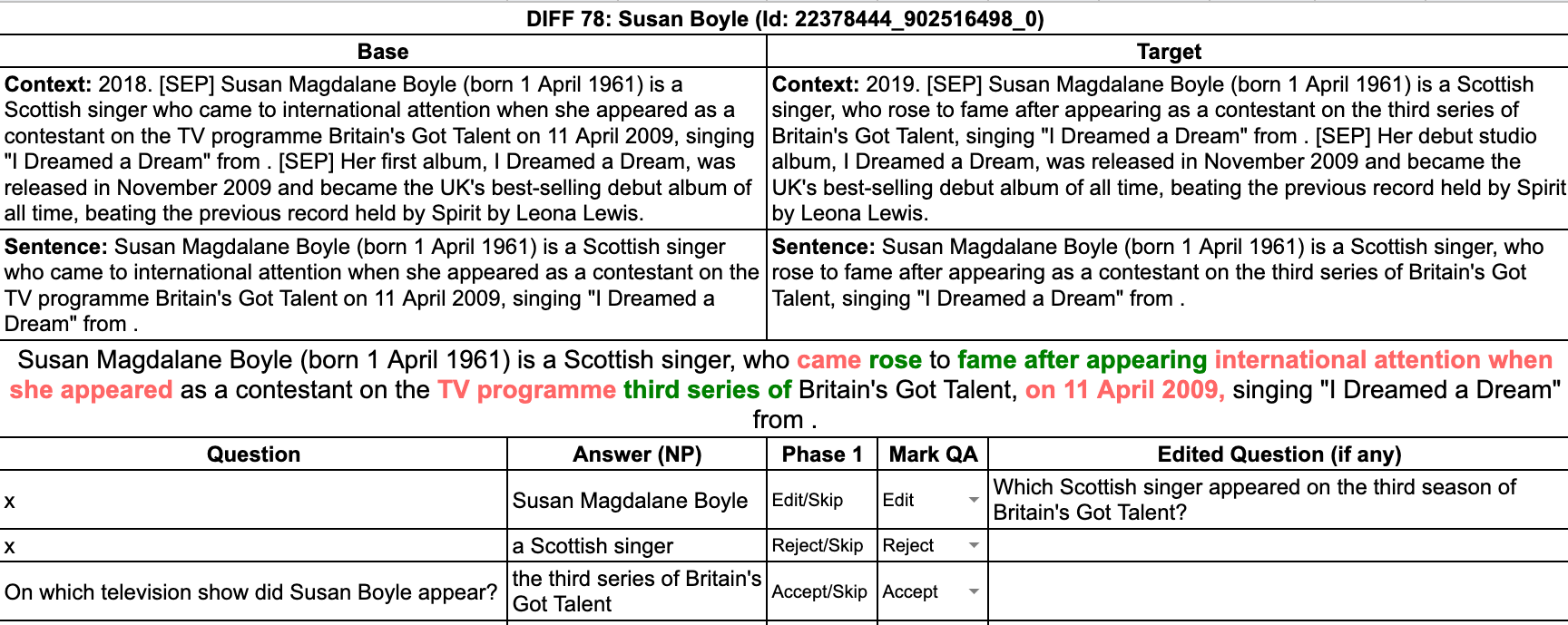}
\caption{Interface for the second phase of annotations, where a third annotator will rephrase a question and/or decide on a disagreed-upon annotation. Here, the annotator writes a new question for the answer span given the \textit{Edit} annotation, and decides to confirm the \textit{Reject} and \textit{Accept} annotations of the other two examples. Note that for \textit{Edit} or \textit{Reject} annotations, to avoid bias, we do not display the seed-question to the annotators and instead display a \textit{x}.}
\label{fig:annotation_p2}
\end{figure*}

Refer \autoref{fig:annotation_p1} and \autoref{fig:annotation_p2} for annotation interface of phase 1 and 2 respectively.


\label{sec:appendix}

\end{document}